# A deep reinforcement learning approach to assess the low-altitude airspace capacity for urban air mobility

**Asal Mehditabrizi**
Graduate Student,
Maryland Transportation Institute, University of Maryland,
20742 College Park, Maryland.
Email: asal97@umd.edu

**Mahdi Samadzad**
Assistant professor,
School of Civil Engineering, College of Engineering, University of Tehran,
1417613131 Tehran, Iran.
Email: msamadzad@ut.ac.ir
* Corresponding Author.

**Sina Sabzekar**
Graduate Student,
Department of Civil Engineering, Sharif University of Technology.
Tehran, Iran. P.O. Box 11365-8639.
Email: sina.sabzekar@sharif.edu

## ABSTRACT

Urban air mobility is the new mode of transportation aiming to provide a fast and secure way of travel by utilizing the low-altitude airspace. This goal cannot be achieved without the implementation of new flight regulations which can assure safe and efficient allocation of flight paths to a large number of vertical takeoff/landing aerial vehicles. Such rules should also allow estimating the effective capacity of the low-altitude airspace for planning purposes. Path planning is a vital subject in urban air mobility which could enable a large number of UAVs to fly simultaneously in the airspace without facing the risk of collision. Since urban air mobility is a novel concept, authorities are still working on the redaction of new flight rules applicable to urban air mobility.

In this study, an autonomous UAV path planning framework is proposed using a deep reinforcement learning approach and a deep deterministic policy gradient algorithm. The objective is to employ a self-trained UAV to reach its destination in the shortest possible time in any arbitrary environment by adjusting its acceleration. It should avoid collisions with any dynamic or static obstacles and avoid entering prior permission zones existing on its path. The reward function is the determinant factor in the training process. Thus, two different reward function compositions are compared and the chosen composition is deployed to train the UAV by coding the RL algorithm in python. Finally, numerical simulations investigated the success rate of UAVs in different scenarios providing an estimate of the effective airspace capacity.

**INTRODUCTION**
Many cities are developing urban air mobility infrastructure and air taxis are scheduled to be flying in the urban airspaces in the coming years [1]. Regular air taxi services are estimated to be profitable in large metropolitan contexts and urban areas are expected to follow as the technology becomes more available [2] .Various methods have been used in previous studies for UAVs' path planning. Most such methods have some weaknesses. For example, in central methods, by adding new information to the system, the program should be rerun which takes a lot of time and energy [3]; mathematical methods need complex calculations, which makes them inappropriate for doing path planning for a large number of UAVs [4] requiring the solution of nonlinear optimization programs [5, 6]; in geometric methods solving the problem become impossible when there is a large number of UAVs in the environment [3]. Deep reinforcement learning is one of the decentralized methods that do not have the weaknesses associated with updating [7]. This approach has been widely used in the domain of autonomous vehicles but its application to the aerial context remains nascent. It is a requirement that appropriate flight rules are developed to coordinate flying in low-altitude airspace [8].

Different reinforcement learning algorithms can be used for UAV path planning. Some of these algorithms, such as Deep Q Network (DQN), have a discrete action space [9] which makes them different from the real world, where UAVs operate according to a continuous action space. Another reinforcement learning algorithm that has widely been used for path planning is the Deep deterministic policy gradient (DDPG). This algorithm is independent of input and output data and can be used for continuous action and state spaces [10]. Thus, it is more realistic than other algorithms.

In path planning using DDPG, UAVs can be trained to follow a be predefined routes with various shapes such as circular or square [11]; or by defining the origin and destination of UAVs, they can define their own routes to reach their destination and avoid having a collision with obstacles in their path [4]. In addition, this algorithm can also be used to control the velocity of UAVs during their path planning [4, 11]. Because of these advantages, we use DDPG reinforcement learning for UAVs' path planning.

A limited number of studies use DDPG with a continuous action space for path planning. Moreover, there are a few studies where UAVs' speed control takes into consideration, and there are no studies where UAVs' acceleration is controlled, while in real-world, UAVs' movement can be controlled by adjusting the acceleration. Therefore, in this study, we use continuous action spaces and control the acceleration of UAVs in addition to their velocity. Furthermore, in this study, we consider prior permission zones (PPZ) and static and dynamic obstacles in our environment. Consequently, unlike previous studies, which usually train and simulate one UAV, we investigate the possibility of simultaneously flying several trained UAVs in the airspace.

**METHOD**
In this section, we describe the characteristics of UAVs, and the environment in which they fly.

**Characteristics of UAVs**
In this study, we should identify UAVs' maximum velocity and acceleration and their flying range, which are essential for their path planning. The characteristics of some UAVs manufactured by different companies are depicted in (**Figure 1**). Based on this information, we assume the maximum velocity of 70 m/s and maximum acceleration of 0.3g, which ensures passengers' comfort [12].

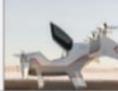

**Figure 1 Characteristics of UAVs [13]**

**Characteristics of the environment**

UAVs' navigation in the real environment is complex. Therefore, we create a virtual environment with a high matching degree to the real-world environment for UAVs' navigation. Unlike most of the previous virtual environments modeled as a grid world, our environment is a free space, and UAVs have the freedom to take any directions to reach their destinations.

We assume that UAVs fly at a constant altitude; thus, our environment would be in 2 dimensions. In this case, since UAVs can only fly in 2 dimensions, their obstacle avoidance would be more challenging. In addition, in the real world, UAVs should fly in a constraint corridor where they do not have much freedom in the 3rd dimension; hence, considering the 2D environment could be reasonable.

We assume that there are static and dynamic obstacles in the environment since UAVs that fly in low-altitude airspace may face obstacles such as tall buildings, flying birds, or other UAVs. The locations of obstacles are random and dynamic obstacles would take the velocity between 20 m/s to 50 m/s and move back and forth in a direction perpendicular to the direction from the origin to the destination of a UAV. The safety distance of the UAVs from the obstacles is considered to be more than 50 meters [14].

We assume that there are prior permission zones (PPZ) in the environment. Flying in these areas is prohibited because of safety or security issues. These zones include hospitals, crowded recreational areas, stadiums, and government buildings [15]. The safety distance of the UAVs from PPZ is considered to be more than 1000 meters.

The origin and destination of each UAV are assumed to be random in each episode. This enables UAVs to perform appropriately in new environments. In addition, we assume that UAVs reach their destination at a 100 meters distance from the exact location point.

**DDPG in UAV path planning**

Deep Deterministic Policy Gradient (DDPG) is an actor-critic reinforcement learning algorithm. This algorithm is model-free and off-policy. Model-free reinforcement algorithms directly optimize value function with trial and error. In these algorithms, we do not need to save environment state in memory so the memory usage is not high. Off-policy algorithms find optimal policy without considering the actor motivation for next action. This policy is independent from value function. The structure of this algorithm is depicted in (**Figure 2**). Variables and factors used in this section are determined in **Table1**.

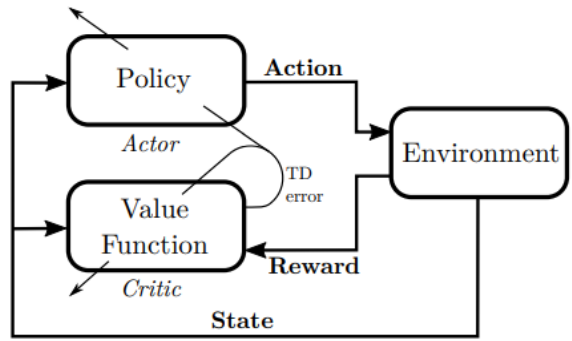

**Figure 2 Actor-critic agent structure [11]**

**TABLE 1 DDPG parameters**

| Symbol | Description | Value |
|---|---|---|
| $\gamma$ | Discount factor for critic updates | 0.9 |
| $\tau$ | Soft target update parameter | 1 |
| $\eta_Q$ | Learning rate of critic network. | 0.0005 |
| $\eta_\mu$ | Learning rate of actor network. | 0.00005 |
| - | Replay buffer size. | 10000000 |
| $t$ | Agent time step. | 1 s |
| - | Maximum steps of one episode | 800 |

The actor chooses the action from the action set which maximizes the value function. This action is calculated based on **Equation 1**.

$$\mu(s_t) = \text{argmax}\, Q(s_t, a_t) \tag{1}$$

The Critic generates the value function based on the environment state and the performed action. The value function is calculated based on the given reward in each step. This function is calculated with Bellman's equation (**Equation 2**).

$$Q(s_t, a_t) = \mathbb{E}_{s_{t+1} \sim E}(r(s_t, a_t) + \gamma \max Q(s_{t+1}, \mu(s_{t+1}))) \tag{2}$$

In this algorithm, actor and critic are neural networks. The steps of this algorithm are as follows:

1. Define actor and critic networks as neural networks with one input layer, one output layer, and two hidden layers with 300 and 400 neurons respectively.
2. Randomly initialize critic and actor neural networks with weights $\theta^Q$ and $\theta^\mu$
3. Initialize target networks with weights $\theta^{Q'} \leftarrow \theta^Q, \theta^{\mu'} \leftarrow \theta^\mu$.
4. Initialize replay buffer b
5. **for** episode = 1, …, M **do**
6.    Receive first observation $s_1$
7.    **for** t = 1, …, T **do**
8.      Select $a_t$ based on ε-greedy algorithm: select random action $a_t$ with ε probability, otherwise choose $a_t$ according to the current policy.
9. Execute action $a_t$ and observe reward $r_t$ and new state $s_{t+1}$.
10. Store transition $[s_t, a_t, r_t, s_{t+1}]$ in b.
11. Sample a random batch of N transitions $[s_j, a_j, r_j, s_{j+1}]$
12. Set $y_j = r_j + \gamma Q'(s_{j+1}, \mu'(s_{j+1}|\theta^{\mu'})|\theta^{Q'})$
13. Update critic by minimizing the loss:
$$L = \frac{1}{N} \sum_j (y_j - Q(s_j, a_j|\theta^Q))^2$$
14. Update the actor policy using policy gradient
15. Update the target networks:
$$\theta^{Q'} \rightarrow \nu \theta^Q + (1 - \tau)\theta^{Q'}$$
$$\theta^{\mu'} \rightarrow \nu \theta^\mu + (1 - \tau)\theta^{\mu'}$$

**Training**

In this section, we train the UAV by a deep reinforcement learning approach and deep deterministic policy gradient algorithm. For this purpose, we use PyTorch library in Python. Three important elements in the training process are states, actions, and reward functions that determine the convergence speed of the algorithm. These elements are different in various environments. Hence, in this paper, we describe them for different environments.

*Environment without obstacles*
      In this case, we train a UAV in an environment without any obstacles or PPZs. We use two different reward functions to compare their performance in training the UAV.

*Mehditabrizi, Samadzad, and Sabzekar*

In this environment, the states, or in other words, the information that the UAV receives, are formed by two states (**Equation 3**): namely, the distance vector from the destination ($d_f$), and the velocity vector ($v$).

$$s_t = \{d_f, v\} \tag{3}$$

The action in this environment is the acceleration vector of the UAV. This acceleration is in x and y directions and can be chosen from a continuous action space as shown in **Equation 4**.

$$\{a_x \in \{-0.3g, 0.3g\}, a_y \in \{-0.3g, 0.3g\}\} \tag{4}$$

The reward function in this study consists of three terms. The first term is a positive reward of reaching a destination ($R_{1_t}$). The second term is a negative reward of quitting the environment ($R_{2_t}$). The last term is a reward of approaching to the destination or moving away from it ($R_{3_t}$) and we use two different functions for this term. The total reward function is shown in **Equation 5.**

$$R_t = T_{reach}\, R_1 + T_{exit}\, R_2 + R_{3_t} \tag{5}$$

In this equation $T_{reach}$ would be 1 if the UAV reach the destination and 0 otherwise. $T_{exit}$ would be 1 if the UAV exit the environment and 0 otherwise.

One of the reward functions that used for the third term is based on the distance from the destination (**Equation 6**).

$$R_{\text{distance}_t} = \beta \exp(-\alpha\, D_{F_t}) - \beta \tag{6}$$

In this equation $D_{F_t}$ is UAV's distance from the destination in each time step and $\beta$ is a constant value. The equation shows that by decreasing the distance from the destination, the reward would increase. Another reward function used for the third term is based on a dot production between the unit vector of distance from the destination ($e_{F_t}$) and the unit displacement vector of a UAV ($e_{d_t}$) (**Equation 7**).

$$R_{\text{dot}_t} = \begin{cases} \alpha(e_{F_t} \cdot e_{d_t}) - \beta & \text{if } (e_{F_t} \cdot e_{d_t}) > 0.9 \\ \gamma(e_{F_t} \cdot e_{d_t}) - \beta & \text{if } (e_{F_t} \cdot e_{d_t}) \leq 0.9 \end{cases} \tag{7}$$

In this equation α, γ and β are constant. If the dot production is more than 0.9, the UAV's path to the destination would be close to the shortest path to the destination.

We train this environment for two reward functions. The success rate of the two models is depicted in (**Figure 3**). It could be seen that by using the dot-product reward function after 2000 episodes the success rate of 100% is achieved. However, by using the distance reward function we achieve a success rate of near 95% after 15000 episodes. Therefore, it can be concluded that the dot-product gives us more learning and convergence speed. Hence, we choose this reward function for training other environments.

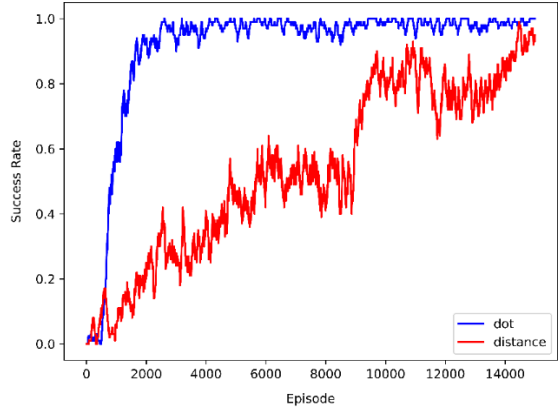

**Figure 3 success rate of agent for two reward functions**

*Environment including obstacles and PPZs*

In this environment, we consider several obstacles and PPZs. The states in this environment include distance vector from the destination ($d_f$), distance vector from the nearest obstacle ($d_O$), distance vector from the nearest PPZ ($d_{PPZ}$), and velocity vector ($v$).

The action in this environment is the acceleration vector of the UAV and can be chosen from **Equation 4**.

The reward function in this study consists of five terms: A positive reward for reaching the destination ($R_{1_t}$), a negative reward for quitting the environment ($R_{2_t}$), a negative reward of colliding the obstacle ($R_{3_t}$), a negative reward for entering the PPZ ($R_{4_t}$), a reward for approaching to or moving away from the destination ($R_{5_t}$) and the obstacle ($R_{6_t}$) and the PPZ ($R_{7_t}$). The reward function can be depicted in **Equation 8.**

$$R_t = T_{reach}\, R_1 + T_{exit}\, R_2 + T_{colission}\, R_3 + T_{PPZ\ entrance}\, R_4 + R_{5_t} + R_{6_t} + R_{7_t} \tag{8}$$

In this equation $R_{5_t}, R_{6_t}, R_{7_t}$ are based on the dot-product reward functions. $R_{5_t}$ increases when the UAV approaches the destination, $R_{6_t}$ increases when UAV moves away from obstacles and $R_{7_t}$ increases when UAV moves away from PPZs.

To train the agent we use the concept of transfer learning. In this machine learning technique, the knowledge is transferred to improve the performance of deep learning models. At first, we train the UAV in an environment with 3 obstacles and 3 PPZs and then we use the trained agent as a base for training in another environment. The new environment has 18 static obstacles, 2 dynamic obstacles and 3 PPZs and it uses to improve the collision avoidance of the UAV when facing static or dynamic obstacles. The success rate (**Figure 4)**, collision rate (**Figure 5**), PPZ entrance rate (**Figure 6)** and exit rate (**Figure 7)** of the UAV are depicted below.

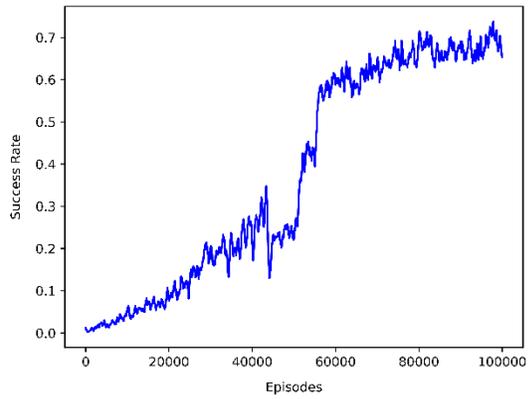
(a)

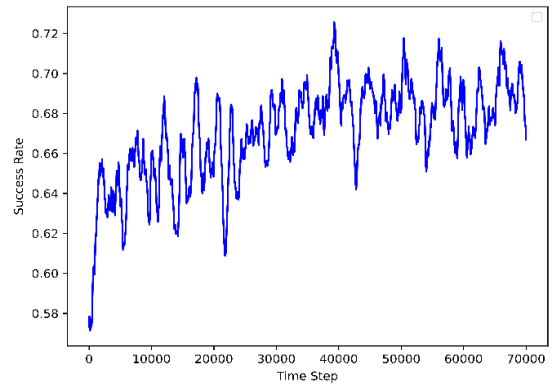
(b)

**Figure 4:** The success rate of the agent in training: (a) the source task (the environment with 3 obstacles and PPZs) and (b) the environment with 18 static obstacles, 2 dynamic obstacles and 3 PPZs)

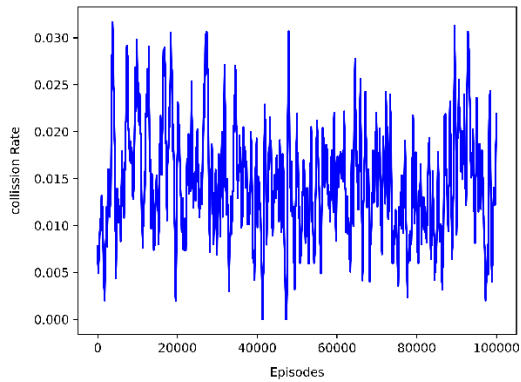
(a)

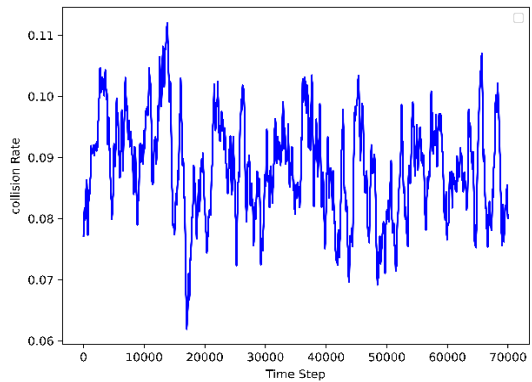
(b)

**Figure 5:** The collision rate of the agent in training: (a) the source task (the environment with 3 obstacles and PPZs) and (b) the environment with 18 static obstacles, 2 dynamic obstacles and 3 PPZs)

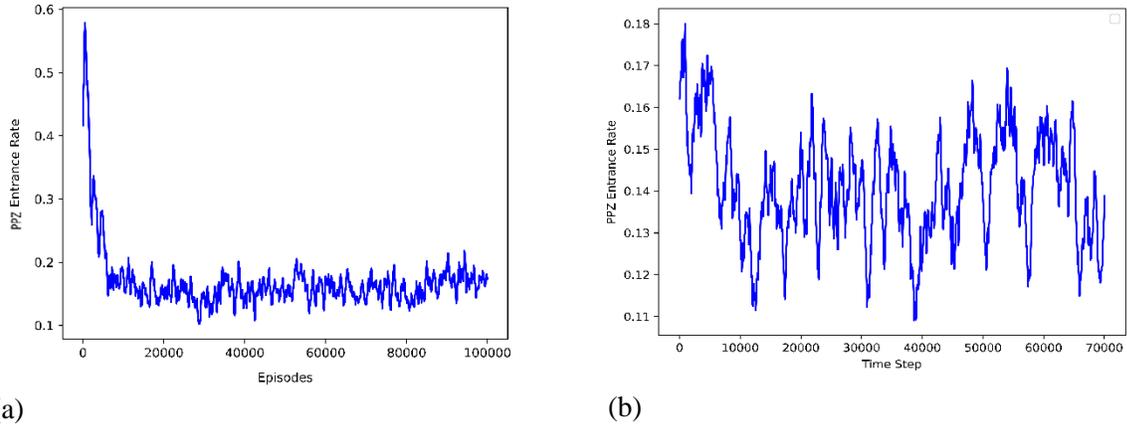

(a)  (b)

**Figure 6:** **The collision rate of the agent in training: (a) the source task (the environment with 3 obstacles and PPZs) and (b) the environment with 18 static obstacles, 2 dynamic obstacles and 3 PPZs)**

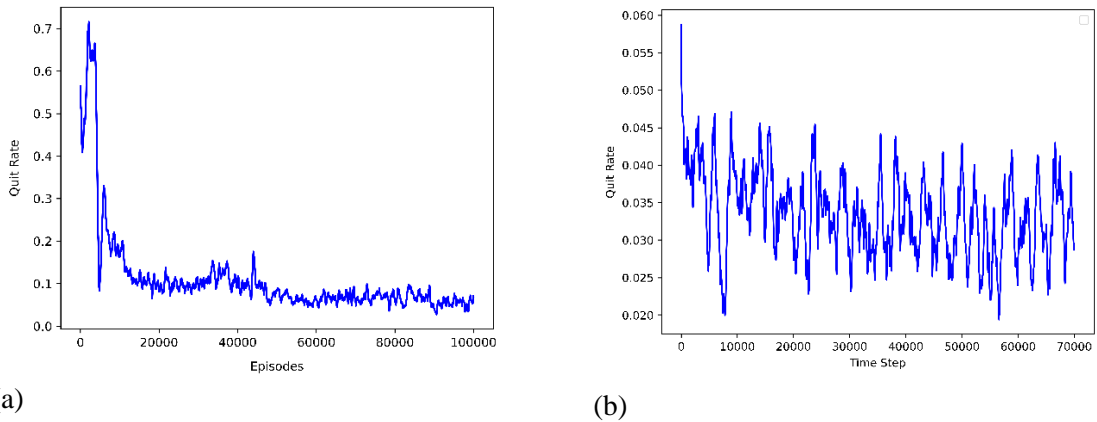

(a)  (b)

**Figure 7:** **The exit rate of the agent in training: (a) the source task (the environment with 3 obstacles and PPZs) and (b) the environment with 18 static obstacles, 2 dynamic obstacles and 3 PPZs)**

**RESULTS**

In this section we study the behavior of the system for selected scenarios. For this purpose, we simulate the trained agent in a specific environment by using Python. In the first scenario we assume that a UAV fly in an environment with one PPZ. The origin and the destination of the UAV are selected randomly and PPZ is between the origin and the destination of UAV. This success rate in this scenario is 100%.
For a specific origin and destination, we draw the UAV's distance from destination and PPZ in each time step (**Figure 8**), and the velocity and acceleration of the UAV in each time step (**Figure 9**).

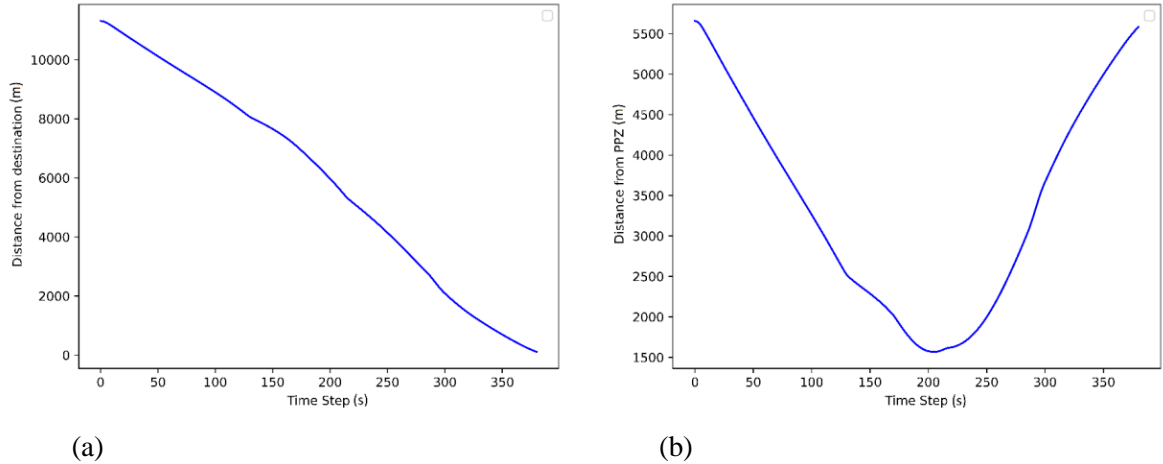

**Figure 8: (a) UAV's distance from destination and (b) UAV's distance from the PPZ**

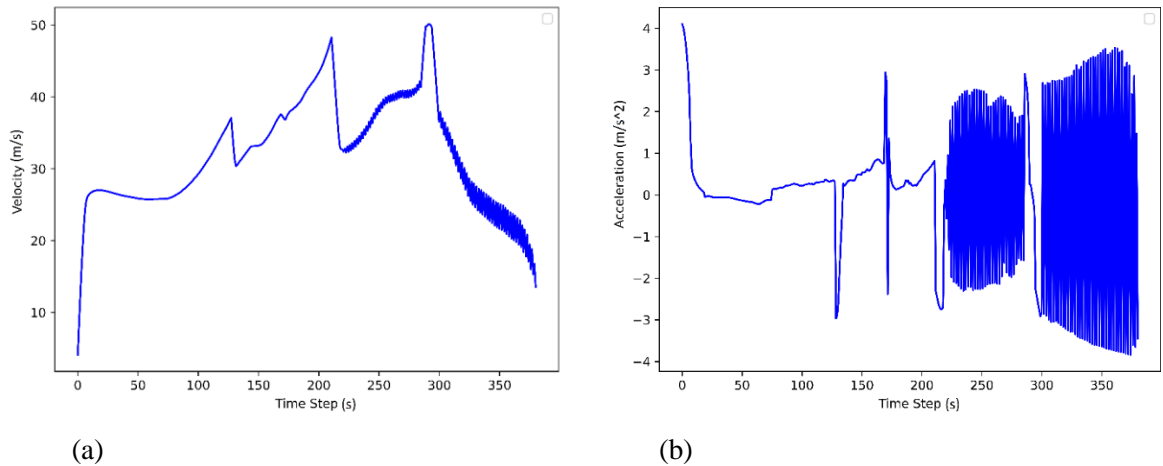

**Figure 9: (a) UAV's velocity and (b) UAV's acceleration**

The two figures show that the UAV reduces its speed significantly near the PPZ to pass it safely. In addition, near the destination, the overall trend of the speed is reducing while it has some fluctuations. This is because, near the destination, the UAV takes the negative acceleration, but its speed might become zero before reaching the destination. Thus, it takes the positive acceleration, and this time it might pass the destination, so it retakes the negative acceleration. This process is repeated until the UAV reaches its destination.

In the second scenario, we assume that there are no PPZs in the environment, and the distance between the origin of UAVs is assumed to be 900 meters. In addition, the origin and destination of UAVs are selected randomly. In this scenario, we estimate the success rate and collision rate of different numbers of UAVs, which are depicted in **Figure 10** and **Figure 11**, respectively.

*Mehditabrizi, Samadzad, and Sabzekar*

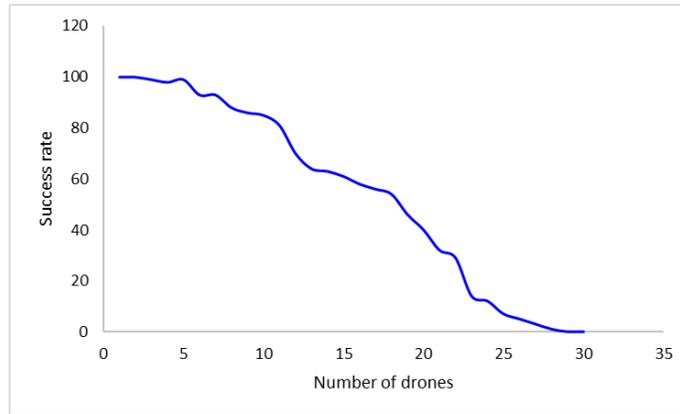

**Figure 10 success rate for different numbers of drones**

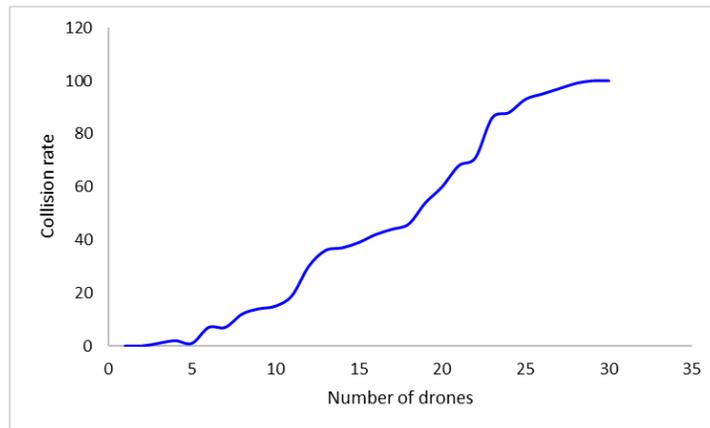

**Figure 11 collision rate for different numbers of drones**

As it could be understood from the above figures, 10 UAVs can fly in a 10km*10km airspace with a success rate near 80%. To understand our model's capability to be implemented in the real world, we investigate the dimensions of some big cities such as Tehran, Paris, and Washington DC. In these cities, there are possible locations for constructing airports for UAVs, and the distance between such locations is about 5 kilometers. Therefore, in a city like Tehran, which is 45 kilometers in one direction and 25 kilometers in the other, the maximum number of UAVs that can fly in one direction simultaneously is 10. This could also be the case in two other cities which are even smaller than Tehran in one direction. Consequently, our model in which 10 UAVs can fly in 10km×10km airspace with a success rate of 80% could be appropriate in the real world in which the airspace is bigger, and there is more distance between the origin and destination of UAVs.

In this scenario, the five UAVs' path for a specific origin and destination is depicted in **Figure 12.** The distance from destination, distance from the nearest obstacle, and the speed of these five UAVs are shown in **Figure 13**. As it could be seen, UAVs reduce their speed near the destination. In addition, when a UAV's distance from another UAV decreases, it reduces its speed to allow safe passage.

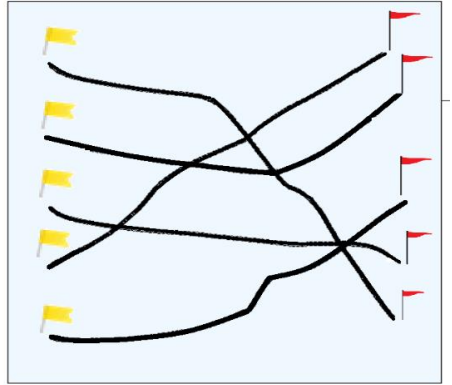

**Figure 12: UAVs' itinerary from origin to destination.**

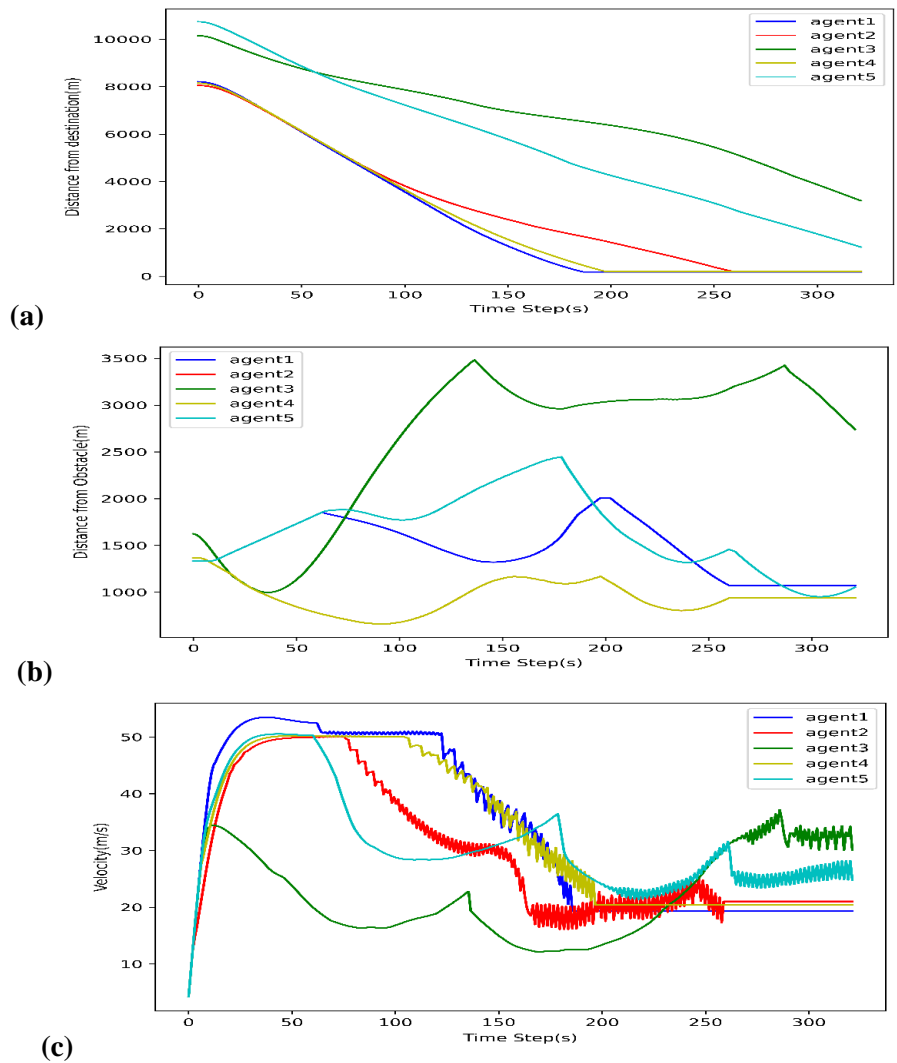

**Figure 13: (a) UAVs' distance from the destination, (b) UAVs' distance from the nearest obstacle, (c) UAVs' speed.**

**CONCLUSIONS**

In this study, we used a DDPG-based deep reinforcement learning approach to do the path planning for UAVs in an environment with PPZs and dynamic and static obstacles. In this path planning, the UAV's acceleration constitutes the action space and represents the throttle of the vehicle, and other cinematic characteristics, such as the speed and traveled distance, are calculated by integration. This provides a more realistic model of the dynamics of the vehicle. The representation of the environment and the action space is continuous.

In this study, we suggested a new approach for attributing values to the reward function based on a dot-product operator and showed that it provides a higher convergence rate compared to the distance-based reward composition.

Finally, by using the proposed method, we could safely navigate several UAVs in different environments with realistic dimensions and obtain an estimate of the success rate as a function of the number of airborne vehicles and restricted areas.

Several directions can be considered for future study some of which include: evaluating the advantage of equipping UAVs with more powerful radar capable of seeing further at a higher sampling rate, representing the third dimension explicitly, and augmenting the reward function with location data communicated from a central control tower.

**AUTHOR CONTRIBUTIONS**
The authors confirm contributions to the paper as follows: study conception and design: A. Mehditabrizi, M. Samadzad; algorithm: A. Mehditabrizi, M. Samadzad; implementation and programming: A. Mehditabrizi, S. Sabzekar; analysis and interpretation of results: A. Mehditabrizi, M. Samadzad, S. Sabzekar; draft manuscript preparation: A. Mehditabrizi, M. Samadzad. All authors reviewed the results and approved the final version of the manuscript.